\pdfoutput=1

\documentclass[11pt]{article}

\usepackage[]{acl}

\usepackage{times}
\usepackage{latexsym}
\usepackage{hyperref}

\usepackage[T1]{fontenc}

\usepackage[utf8]{inputenc}

\usepackage{microtype}

\let\oldFootnote\footnote
\newcommand\nextToken\relax

\renewcommand\footnote[1]{%
    \oldFootnote{#1}\futurelet\nextToken\isFootnote}

\newcommand\isFootnote{%
    \ifx\footnote\nextToken\textsuperscript{,}\fi}

\usepackage{wrapfig,lipsum,booktabs}
\usepackage{latexsym}
\usepackage{amssymb}
\usepackage{times}

\usepackage{soul}
\usepackage{url}

\usepackage[utf8]{inputenc}
\usepackage{graphicx}
\usepackage{amsmath}
\usepackage{booktabs}
\usepackage{microtype}

\usepackage{arydshln}
\RequirePackage{xcolor}
\definecolor{darkgreen}{RGB}{0,119,0} 

\newcommand{\convirt}{\textsf{MM-SimCLR}}
\newcommand{\pienet}{\textsf{Ext-PIE-Net}}
\newcommand{\memotion}{\textsf{Memotion}}
\newcommand{\harmeme}{\textsf{HarMeme}}
\newcommand{\harmp}{\textsf{Harm-P}}
\newcommand{\mmhs}{\textsf{MMHS150K}}
\newcommand{\facebook}{\textsf{Hateful Memes}}
\newcommand{\sslft}{\textsf{SSL+FT}}
\newcommand{\finetune}{\textsf{FT}}
\newcommand{\selfsupervised}{\textsf{SSL}}
\newcommand{\supervised}{\textsf{SL}}
\newcommand{\sent}{\textsf{SENT}}
\newcommand{\emot}{\textsf{EMOT}}
\newcommand{\emotq}{\textsf{EMOT-Q}}

\usepackage{mathrsfs}
\usepackage{bbm}

\usepackage{blindtext}

\usepackage{subcaption}
\usepackage{xcolor}
\usepackage{multirow}
\usepackage[inline]{enumitem}
\usepackage{soul}
\usepackage{xcolor}

\title{Domain-aware Self-supervised Pre-training \\for Label-Efficient Meme Analysis}

\author{Shivam Sharma$^{1,4}$, Mohd Khizir Siddiqui$^{3}$, Md. Shad Akhtar$^1$ and Tanmoy Chakraborty$^2$\\
  $^1$Indraprastha Institute of Information Technology Delhi, India  \\
  $^2$Indian Institute of Technology Delhi, India  \\
  $^3$Birla Institute of Technology and Science, Goa, India \\
  $^4$Wipro AI Labs, India\\
  \small\texttt{\{shivams, shad.akhtar\}@iiitd.ac.in}, \small\texttt{mdkhizirsiddiqui@gmail.com}, \small\texttt{tanchak@ee.iitd.ac.in}}

\begin{document}
\maketitle
\begin{abstract}
Existing self-supervised learning strategies are constrained to either a limited set of objectives or generic downstream tasks that predominantly target uni-modal applications. This has isolated progress for imperative multi-modal applications that are diverse in terms of complexity and domain-affinity, such as meme analysis. Here, we introduce two self-supervised pre-training methods, namely \pienet\ and \convirt\ that (i) employ off-the-shelf multi-modal hate-speech data during pre-training and (ii) perform self-supervised learning by incorporating multiple specialized pretext tasks, effectively catering to the required complex multi-modal representation learning for meme analysis. 

We experiment with different self-supervision strategies, including potential variants that could help learn rich cross-modality representations and evaluate using popular linear probing on the \facebook\ task. The proposed solutions strongly compete with the fully supervised baseline via label-efficient training
while distinctly outperforming them on all three tasks of the \memotion\ challenge with $0.18\%$, $23.64\%$, and $0.93\%$ performance gain, respectively. Further, we demonstrate the generalizability of the proposed solutions by reporting competitive performance on the \harmeme\ task. Finally, we empirically establish the quality of the learned representations by analyzing task-specific learning, using fewer labeled training samples, and arguing that the complexity of the self-supervision strategy and downstream task at hand are correlated. Our efforts highlight the requirement of better multi-modal self-supervision methods involving specialized pretext tasks for efficient fine-tuning and generalizable performance.
\end{abstract}

\section{Introduction}

The overwhelming scale of digital mutation constantly transpiring over the web is ``creating the illusion of reality, addressing the viewer, and representing a convoluted space" \cite{manovich2001language}. Almost every social activity affects or is affected by an online entity, sometimes even disturbing social harmony, influenced by a prominent surge of multi-modal harmful, abusive and hateful online content. Therefore, it is imperative to explore solutions towards automatic mediation of online activities that pre-dominantly involve multi-modality. Recently, there has been a defining resurgence of advancements in multi-modal AI, albeit slowly. 

Existing self-supervision strategies for visual-linguistic applications involve different \textit{pretext} tasks like Masked Language Modeling (MLM) \cite{Devlin2019BERTPO}, Masked Region Modeling (MRM) \cite{chen2020uniter}, Word-Region Alignment (WRA) \cite{Gupta2017AlignedIR}, and Image-Text Matching (ITM) \cite{Li2019VisualSR,radford2021learning}, which inherently presume visual-linguistic grounding \cite{karpathy_vissem_2015}. As a consequence, the large-scale datasets like MS COCO \cite{Lin2014MicrosoftCC}, Conceptual Captions (CC) \cite{sharma2018conceptual}, Wikipedia-based Image Text (WIT) \cite{wit_Srinivasan} and LAION-400M \cite{Birhane2021}, curated towards the required pre-training, are either mostly generic in nature or represent a greater degree of visual-semantic association between the image and text pairs. Moreover, the required multi-modal datasets are rather challenging to create, as they often require multi-dimensional and fine-grained manual annotations for a large volume of multi-modal data. 

These frameworks have demonstrated impressive pre-training schemes for addressing downstream multi-modal tasks like Visual Question Answering (VQA), Image Captioning (IC), Visual Commonsense Reasoning (VCR), etc. \cite{CVNLP}. Still, there is significant room for improvement in terms of their generalizability. For instance, besides \textit{masked language modelling} (MLM), state-of-the-art multi-modal models like Visual BERT, ViLBERT and LXMERT are pre-trained wrt pretext tasks like  \textit{sentence-image prediction} \cite{li2019visualbert}, \textit{masked multi-modal learning, multi-modal alignment prediction} \cite{Lu2019ViLBERTPT} and \textit{detected-label classification} \cite{tan-bansal-2019-lxmert}, which presume aspects like availability of multiple \textit{semantically grounded} sentences corresponding to an image and visual-semantic object and pixel-level annotations for the images.  These requirements constrain modeling aspects for multi-modal content like \textit{memes}. Although such approaches address the issue of scale and cross-modal alignment in terms of \textit{common-sense} reasoning extremely well, they tend to fall short on performance for complex multi-modal tasks like meme analysis \cite{chen2020simple,kiela2020hateful}. This is because memes \textit{do not} represent strong visual-linguistic grounding and solicit sophisticated multi-modal fusion along with contextual knowledge integration.


This paper presents the design and evaluation of efficient multi-modal frameworks that do not rely upon large-scale dataset curation and annotation and can be pre-trained using the datasets from the wild. Also, the pre-training employed is optimally designed toward learning enriched multi-modal representations, which can be further used for addressing downstream tasks like meme analysis in a label-efficient manner. Our contributions, as enlisted below, are three-fold:
\begin{enumerate}[leftmargin=*]
\setlength{\itemsep}{0pt}
    \item We propose two self-supervision-based multi-modal pre-training frameworks which learn semantically rich cross-modal features for meme analysis.
    \item We empirically establish the efficacy of the proposed self-supervision frameworks towards adapting to downstream tasks using only a few labeled training samples.
    \item We finally demonstrate the generalizability of the representations learned across tasks and datasets.\footnote{The source codes are uploaded as supplementary material.}
\end{enumerate}

\section{Related Work}
\textbf{Self-supervised and Semi-supervised Learning: } 
Self-supervised learning approaches are formulated to optimize training objectives that do not require an explicit set of labels. They incorporate pretext tasks to introduce pseudo-labels and learn embedding space rather than solving a specific downstream task. One of the prominent pretext tasks for pre-training language models is next word prediction using a part of the sentence \cite{Peters2018DeepCW}. 
ALBERT \cite{Lan2020ALBERTAL} performs sentence order prediction (SOP) to achieve a similar objective.

Although self-supervision has taken long strides for NLP applications, it has taken a while to show promise for vision applications. A prominent series of work aims at optimizing the similarity between positive pairs of augmented representations while reducing it for negative pairs \cite{oord2018representation}, \cite{chen2020simple}, 
also known as contrastive learning. A non-contrastive learning approach increases similarity with the previous versions of augmented views \cite{grill2020bootstrap}. Such works have long been attempting to solve problems about specific modalities only. We aim to learn multi-modal embedding space enriched to solve non-trivial downstream tasks.

\noindent \textbf{Multi-modal Pre-training: }Recently, \citet{wang2021simvlm} proposed a simple yet effective multi-modal system with specialized convolution layers at the beginning of the encoder and a textual decoder as a follow-up. Other recent similar works include DALL-E \cite{ramesh2021zeroshot}, a zero-shot, generative scalable Transformer that models multi-modal information in an auto-regressive manner and is conditioned on a textual query. This is followed by CLIP, a contrastive learning-based model \cite{radford2021learning}, which is pre-trained on $400$ million image-text pairs collected from different web-based resources. The primary objective of such efforts is to learn multi-modal embedding space jointly. However, the datasets used to pre-train are too generic 
to capture complex semantics.
In this work, we intend to examine such constraints and their impact on the performance of multi-modal systems.  
%

\noindent \textbf{Studies on Memes: } Although the recent past has witnessed an overwhelming amount of research related to memes, especially for topics like online hate, harm, offense, abuse, etc. \cite{kiela2020hateful,sharma-etal-2020-semeval}, still, there are a wide array of meme related tasks, that are yet to be addressed. \citet{Kolwole} explored the classification task on a small dataset and with a linear SVM on low-level descriptors, leveraging only visual information.
Significant efforts have been invested towards meme generation by representing the meme image and the catchphrase in the same vector space using a deep neural network \cite{kido}, leveraging pre-trained Inception-v3 network-based feature extraction. This was further explored in \cite{Dank} for caption generation and rule-based classification. The human assessment in this study outperformed random choices. The quality, however, was below-par as compared to human-produced memes. Efforts are solicited wherein richer and more meaningful content modeling is achieved towards solving tasks that conventional multi-modal approaches cannot.

\section{Dataset}
\paragraph{Pretraining:} To address generalizability towards an array of such topics, we employ the \mmhs\ dataset \cite{Gomez2020ExploringHS} as our primary data source for pre-training our proposed systems. It consists of $150$K multi-modal (images + text) tweets spanning over four hate-inclined topics -- \textit{racism}, \textit{sexism}, \textit{homophobia}, and \textit{religious extremism}. Moreover, the images in the dataset represent diversity with the presence of memes, morphed images, satirical art, etc.

Besides this, to ensure that our pre-training dataset reasonably represents the content type we would evaluate as part of downstream tasks, we also add the memes from the training split of the Facebook's \facebook\ dataset \cite{kiela2020hateful}, that we reserve exclusively for our pre-training. 

\paragraph{Training and Evaluation:} We employ three datasets (\facebook, \harmp, and \memotion) and five different tasks (\textit{hate detection}, \textit{harmfulness detection}, \textit{sentiment analysis}, \textit{emotion classification}, and \textit{emotion class quantification}) to demonstrate the efficacy of our proposed approaches. The \harmp\ dataset belongs to the \harmeme\ task \cite{pramanick-etal-2021-momenta-multimodal} and consists of 3552 memes annotated with two labels -- \textit{harmful or not-harmful}. The \memotion\ dataset \cite{sharma-etal-2020-semeval} has approx. 8K memes and defines three subtasks\footnote{We use abbreviations \sent,\ \emot\ and \emotq\ for \textit{sentiment analysis}, \textit{emotion classification}, and \textit{emotion class quantification}, respectively.} -- \textit{sentiment analysis} (positive/negative), \textit{emotion classification} (humour/sarcasm/offense/motivational), and \textit{emotion class quantification} (slightly/mildly/very). Although these datasets are based on memes or multi-modal content, their objectives are different and have \textit{varying} complexities.
\footnote{We present further details like lexical characteristics and text-length comparison for the datasets used in App. \ref{app:datanalysis}.}.

We leverage a dataset that represents the raw, unprocessed large-scale corpus of multi-modal information, specifically emphasizing different types of hate speech. We acknowledge that a labeling scheme initially accompanies the dataset (MMHS150K). However, we do not utilize that information either during the pre-training stage or during the task-specific fine-tuning stage. This is also represented in the form of proposed loss functions, which do not utilize source data labels but solely rely on the intermediate neural representations, hence self-supervised. Also, the underlying presumption for utilizing such a dataset (MMHS150K) in a self-supervised way is based on the fact that the original dataset owners collected it using a pre-defined set of database keywords \cite{Gomez2020ExploringHS}, and this is all that one would need to do to obtain such a dataset at scale towards pre-training the models proposed. Also, no explicit annotation process is required for pre-training \convirt\ and \pienet. Now, as for the task-specificity, we already showcase the performances of the fully supervised systems that utilize fine-tuning of the models, pre-trained using a generic dataset. We propose the frameworks that, if pre-trained using a "domain-oriented" dataset that can be easily obtained, without any special annotations, can quickly and in a label-efficient way adapt to related downstream tasks.


\section{Proposed Solution}

We propose two methods: \convirt\ and \pienet,\ that utilize adaptations of popular contrastive and triplet loss formulations for learning multi-modal embedding space. Proposed solutions also encapsulate specialized multi-modal pretext tasks suited toward joint multi-modal representation learning. Before describing the proposed solutions, we first review the two-loss formulations below.


\noindent $\bullet$ \textit{\underline{SimCLR}:} The SimCLR framework \cite{chen2020simple}, a popular self-supervision technique, learns representations for images by maximizing agreement between their augmented views in a latent space. The objective function is defined as:

\begin{small}
\setlength\abovedisplayskip{0pt}
\begin{equation}
    \mathcal{L}_{(i,j)}^{\text{NT-Xent}} = - \log \frac{\exp(\text{sim}(\mathbf{z}_{i}, \mathbf{z}_{j}) / \tau)}{\sum_{k=1}^{2N} \mathbbm{1}_{[k \neq i]} \exp(\text{sim}(\mathbf{z}_{i}, \mathbf{z}_{k}) / \tau)}
    \label{eq:simclr}
\end{equation}
\end{small}
where $\mathbbm{1}_{[k \neq i]}\in\{0,1\}$ is an indicator function; 
 $\mathbf{z}_{i}$ and $\mathbf{z}_{j}$ are the projections for augmented views $i$ and $j$, respectively; and $\tau$ is temperature. 

\noindent $\bullet$ \textit{\underline{Hinge Loss}:} Conventionally, hinge loss has been known to be applied to characterize optimization in uni-modal vector space \cite{Rosasco_areloss}. The formulation of the multi-modal hinge loss has been employed in \cite{Faghri2018VSEIV}.
For a two-modality system with $u$ and $v$ as modality-specific representations in common space, a multi-modal weighted hinge loss ($\mathcal{L}^{\text{wHinge}}$) is formulated using a cosine similarity function $s(\cdot)$. It assumes a margin of $\alpha$ and clamps the value with a ReLU function. 
Moreover, the individual terms are weighted by $\lambda_{u2v}$ and $\lambda_{v2u}$ before aggregation. This is expressed as follows:

\begin{small}
\setlength\abovedisplayskip{0pt}
\begin{multline}
    \mathcal{L}^{\text{wHinge}}(\textbf{u}, \textbf{v}) = 
    \lambda_{u2v} \sum_{\widehat{u}} \text{ReLU} \Big{(} \alpha - s(\textbf{u}, \textbf{v}\>) + s(\widehat{\textbf{u}}, \textbf{v}\>) \Big{)} \\
    +  \lambda_{v2u}  \sum_{\widehat{v}} \text{ReLU} \Big{(} \alpha - s(\textbf{u}, \textbf{v}\>) + s(\textbf{u}, \widehat{\textbf{v}}\>) \Big{)}
    \label{eq:whinge}
\end{multline}
\setlength{\belowdisplayskip}{0pt}
\end{small}

\begin{figure}[t!]
\centering
\resizebox{\columnwidth}{!}{%
\includegraphics[width=\columnwidth]{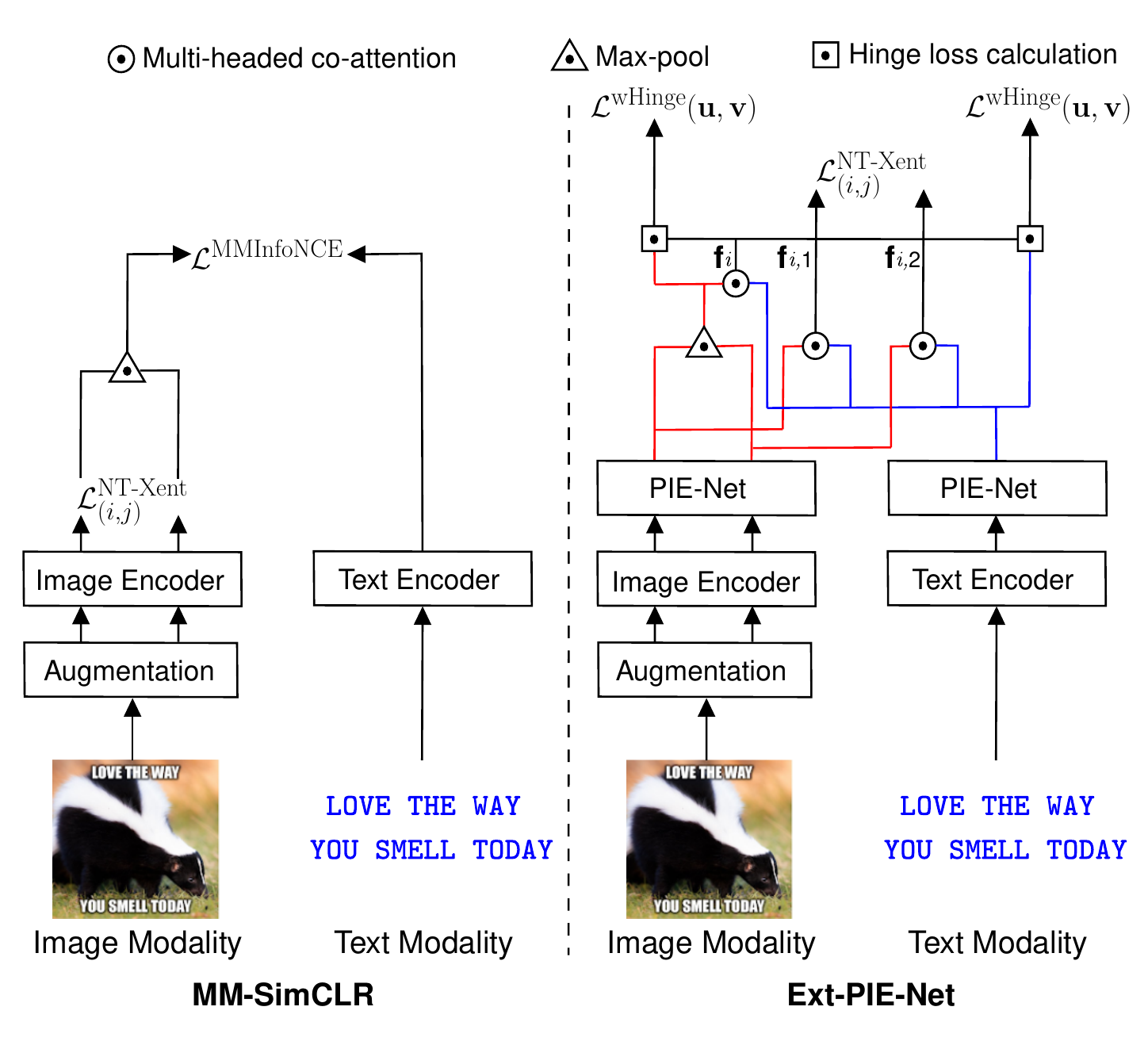}}
\caption{
    Solution architectures of multi-modal self-supervision for memes. \convirt: Multi-modal SimCLR (left); \pienet: Extended Pie-Net (right).
}
\label{fig:ssl_ft_arch}
\vspace{-5mm}
\end{figure}

\paragraph{\convirt:} 
In our first approach, \convirt,\ we integrate discriminative modeling capacity, which leverages contrastive learning in the latent space for images and a dedicated formulation for a multi-modal setup. This is motivated by \citep{zhang2020contrastive}, which performs contrastive learning between the medical images and their associated texts.
Their objective function $\mathcal{L}$ constitutes two terms ($\ell_{i}^{u \rightarrow v}$ and $\ell_i^{v \rightarrow u}$) to maximize association between image and text representations ($\textbf{u}_i$ and $\textbf{v}_i$). Both $\textbf{u}_i$ and $\textbf{v}_i$ are normalized to unit-vectors before being incorporated into the loss terms. $\tau$ is a scaling factor that controls the sensitivity of association, and $\lambda$ controls the weight of the individual term in the final equation. This is given by:

\begin{small}
\setlength\abovedisplayskip{0pt}
    \begin{equation}
        \ell_i^{v \rightarrow u} = - \log \frac{\exp(\langle\textbf{v}_i, \textbf{u}_i \rangle / \tau)}{\sum_{k=1}^N\exp(\langle\textbf{v}_i, \textbf{u}_k \rangle / \tau)}
    \end{equation}
    \begin{equation}
        \ell_i^{u \rightarrow v} = - \log \frac{\exp(\langle\textbf{u}_i, \textbf{v}_i \rangle / \tau)}{\sum_{k=1}^N\exp(\langle\textbf{u}_i, \textbf{v}_k \rangle / \tau)}
    \end{equation}
\setlength{\belowdisplayskip}{0pt}
\end{small}

We will refer to this objective function as Multi-modal InfoNCE loss in our work, given by:

\begin{small}
\setlength\abovedisplayskip{0pt}
    \begin{equation}
        \mathcal{L}^{\text{MMInfoNCE}} = - \frac{1}{N} \sum_{i=1}^N(\lambda \ell_{i}^{u \rightarrow v} + (1-\lambda)\ell_{i}^{v \rightarrow u})
        \label{eq:mminfonce}
    \end{equation}
\end{small}

Finally, we formulate a new objective function for \convirt\ as the summation of SimCLR (Eq. \ref{eq:simclr}) and Multi-modal InfoNCE (Eq. \ref{eq:mminfonce}) losses. The overall process flow is shown in Fig. \ref{fig:ssl_ft_arch} (left).

\begin{small}
\setlength\abovedisplayskip{0pt}
    \begin{equation}
        \label{mmloss}
        \mathcal{L} = \mathcal{L}^{\text{MMInfoNCE}} + \sum_{i=1}^{N} \mathcal{L}^{\text{NT-Xent}}_i
    \end{equation}
\setlength{\belowdisplayskip}{0pt}
\end{small}
\paragraph{\pienet:} 
Inspired by PIE-Net \cite{song2019polysemous}, which is a diversity-inducing visual-semantic embedding learning framework, 
we propose \pienet, which optimizes an \textit{augmented} multi-modal objective function (in Eq. \ref{eqn:loss/pienet}). PIE-Net leverages a representation learning scheme to cater to the lexical diversity within languages via symmetric cross-modal loss formulations. On the other hand, we augment such a formulation by factoring in an additional loss term due to image-specific contrastive loss. It essentially has three major components -- SimCLR $\mathcal{L}^{\text{NT-Xent}}$ (Eq. \ref{eq:simclr}) and a pair of weighted hinge losses $\mathcal{L}^{\text{wHinge}}$ (Eq. \ref{eq:whinge}). 
$\mathcal{L}^{\text{NT-Xent}}$ optimizes the agreement between the augmented multi-modal representations $\textbf{f}_{i, 1}$ and $\textbf{f}_{i, 2}$. We compute these multi-modal representations using multi-headed co-attention between the textual and visual representations. The intuition is to leverage the contrasting representations of the visual and textual modalities. 

We then fuse image views via max-pooling and subsequently with the textual representation using multi-headed co-attention. The obtained multi-modal representation helps in computing modality-reinforcing weighted hinge losses, $\mathcal{L}^{\text{wHinge}}(\textbf{i}_i, \textbf{f}_i)$ and $\mathcal{L}^{\text{wHinge}}(\textbf{t}_i, \textbf{f}_i)$, \textit{w.r.t.} the image ($\textbf{i}_i$) and text ($\textbf{t}_i$) representations, respectively.
The losses are weighted by $\lambda_{f2f}$ ($=0.6$), $\lambda_{f2i}$ ($=0.2$) and $\lambda_{f2t}$ ($=0.2$) to compute the final loss $\mathcal{L}$. Fig. \ref{fig:ssl_ft_arch} (right) shows the \pienet\ framework.

\begin{small}
\setlength{\abovedisplayskip}{0pt}
\begin{multline}
    \label{eqn:loss/pienet}
    \mathcal{L} = \sum_i^N \Big{[} \lambda_{f2f} \cdot \mathcal{L}^{\text{NT-Xent}}(\textbf{f}_{i, 1}, \textbf{f}_{i, 2})
    + \lambda_{f2i} \cdot \mathcal{L}^{\text{wHinge}}(\textbf{i}_i, \textbf{f}_i)\\
    + \lambda_{f2t} \cdot \mathcal{L}^{\text{wHinge}}(\textbf{t}_i, \textbf{f}_i) \Big{]}
\end{multline}
\end{small}
\section{Experiments and Results}
This section presents the evaluation strategy, description of systems examined, results of experiments on self-supervision, and downstream evaluation. We first experiment with various self-supervision strategies and then evaluate the representations learned from best-performing systems by evaluating different downstream tasks for label-efficient supervised learning.\footnote{We use abbreviations \supervised,\ \selfsupervised\ and \finetune\ for supervised, self-supervised learning, and fine-tuning, respectively.}\textsuperscript{,}\footnote{\label{note:appen}Additional details of experiments, along-with hyperparameters explored are included as part of App. \ref{sec:hyper}.}

To evaluate the representations learned through pre-training, we employ the linear evaluation strategy \cite{oord2018representation}, which trains a linear classifier with frozen base network parameters. This is a popular strategy for assessing the quality of the representations learned with a minimal predictive modeling setup that facilitates a fair assessment of the resulting inductive bias. The performance on the test set implies the quality of the representations learned. Since the primary focus of our work is self-supervision for multi-modal applications, we emphasize our investigation and compare mainly with the multi-modal state-of-the-art setups. Also, as we motivate in the Introduction section, standardized large-scale multi-modal datasets like MS-COCO, CC, etc., used towards pre-training visual-linguistic models like ViLBERT \cite{Lu2019ViLBERTPT} and Visual BERT \cite{li2019visualbert} incur significant development cost, we mostly restrict our \sslft\ comparison either to the setups that can conveniently leverage raw datasets like MMHS150K \cite{Gomez2020ExploringHS}, which are conveniently accessible via web (\textit{one of the primary motivations for this work}), or pre-trained and fine-tuned versions of ViLBERT and Visual BERT. 
For comparison, we comply with the respective works and compute accuracy values for the \facebook\ task and Macro-F1 scores for the \memotion\ and \harmeme\ tasks and report all the results by taking the average across \textit{five} independent runs. 

\subsection{Self-supervised Learning and Linear Evaluation}
\paragraph{Systems:}
We experiment with a few existing related approaches and different uni-modal and multi-modal variants and compare self-supervised and supervised learning frameworks for a comprehensive assessment. 
We do not consider explicit pre-training of models like Visual BERT and ViLBERT within the scope of the current study because their pre-training strategies are designed for explicitly modeling visual-linguistic grounding. This can constrain the self-supervised learning based upon \textit{domain-aware} pre-training, using a dataset from the wild (WWW), which is a crucial aspect of our study. However, we do compare the \sslft\ systems with completely fine-tuned and pre-trained checkpoints of Visual BERT (MS-COCO) and ViLBERT (CC) systems. The details of these systems are enlisted as follows:
    $\bullet$ SimCLR \cite{chen2020simple}: The framework focuses on incentivizing the agreement between similar image views.
    $\bullet$ VSE++ \cite{Faghri2018VSEIV}: It focuses on mining hard negatives to heavily penalize for dissimilarity with the anchor images through a hinge-like loss. 
    $\bullet$ Modified SimCLR: We try to extend the loss proposed in SimCLR to text modality via augmentation. We do so using WordNet \cite{wordnet} synonyms replacement and through back-translation \cite{sennrich-etal-2016-improving} approaches. 
    

We also compare state-of-the-art multi-modal systems for better task-specific assessment. These are: 
$\bullet$ Late fusion: Averages prediction scores of ResNet-152 and BERT.
    $\bullet$ Concat BERT: Concatenates representations from ResNet-152 and BERT, using a perceptron as a classifier.
    $\bullet$ MMBT: Multimodal Bitransformer \cite{kiela2020supervised}, capturing the intra/inter-modal dynamics.
    $\bullet$ ViLBERT CC: Vision and Language BERT \cite{lu2019vilbert}, trained on an intermediate multi-modal objective (conceptual captions) \cite{sharma2018conceptual}, comprises of task-independent joint representation multi-modal framework.
    $\bullet$ Visual BERT COCO: Pre-trained \citep{li2019visualbert} using MS-COCO dataset \citep{Lin2014MicrosoftCC}.

\begin{table}[t!]
\centering
\resizebox{0.75\columnwidth}{!}
{
\begin{tabular}{c|l|c}
\hline
\textbf{Type} & \multicolumn{1}{c|}{\textbf{Model}} & \textbf{Acc.} \\ \hline
\multirow{6}{*}{\supervised} & Image-Grid (image-only) & 0.507\\ \cdashline{2-3}
& ViLBERT & 0.631 \\ \cdashline{2-3}
& ViLBERT CC & 0.661 \\ \cdashline{2-3}
& Visual BERT & 0.650 \\ \cdashline{2-3}
 & Visual BERT COCO & 0.659 \\ \cdashline{2-3}
 & alfred lab  & \textbf{0.732} \\ \hline\hline
 \multirow{6}{*}{\selfsupervised} & SimCLR (image-only) & 0.500\\ \cdashline{2-3}
 & Mod. SimCLR-WN & 0.481\\ \cdashline{2-3}
 & Mod. SimCLR-BT & 0.450\\ \cdashline{2-3}
 & VSE  & 0.501 \\ \cdashline{2-3}
 & VSE++$^{\dagger}$  & 0.536 \\ \cline{2-3}
 & \convirt & 0.551 \\ \cdashline{2-3}
 & \pienet$^{\star}$ & \textbf{0.600} \\\hline\hline
\multicolumn{2}{c}{$\Delta_{\text{($\star$-$\dagger$)}\times 100}(\%)$} & \textcolor{blue}{$\uparrow6.42\%$}\\\hline
\end{tabular}
}
\caption{Comparison between the proposed \selfsupervised\ method and baselines on the \facebook\ dataset. $\dagger$ represents \selfsupervised\ baseline and $\star$ is for the proposed approach.}
\label{tab:res_ssl_base}
\end{table}

\paragraph{Results:}
We first examine representations learnt by SimCLR \cite{chen2020simple} and evaluate them by fine-tuning on \facebook\ task. As shown in Table \ref{tab:res_ssl_base}, this results in a meagre accuracy of $0.50$ -- a difference of only $0.67\%$ against the image-only \textit{fully supervised} baseline (accuracy $0.5067$). Moving forward, our initial attempt toward modeling multi-modality involves evaluating a VSE++ \cite{Faghri2018VSEIV} setup, which leverages \textit{hard-negative} sampling to distinguish similar and dissimilar representations. Due to the factoring of hard-negatives in VSE++, the mutual information between the representations of semantically close image-text pairs is regulated and yields an improved accuracy of $0.53$.
Our attempt to extend SimCLR for textual modality results in low accuracy values of $0.48$ and $0.45$, respectively. The low performances are possible due to the changes in the textual semantics that augmentation techniques could induce, effectively reducing potential harmfulness modeling affinity. 

In comparison, \convirt\ enhances the performance, yielding an accuracy of $0.5508$. \pienet\ is observed to further enhance it to $0.5998$ -- a gain of $+9.98\%$ over the image-only SimCLR framework, whereas $+9.84\%$ and $+6.42\%$ over the multi-modal VSE and VSE++ systems respectively (Table \ref{tab:res_ssl_base}). One of the characteristic changes that the proposed solutions incorporate in contrast to the other frameworks is the combined consideration of multiple image views and a single textual representation toward modeling a specialized multi-modal contrastive learning setup. This is likely responsible for the cross-modal efficacy observed in the performance. Although the performances of the proposed models fall behind that of their fully-supervised counterparts, they perform reasonably better than the strong self-supervised methods.


\begin{table}[t!]
\centering
\resizebox{\columnwidth}{!}
{
\begin{tabular}{c|l|c|c|c}
\hline
\multirow{2}{*}{\textbf{Type}}                                                                                   & \multicolumn{1}{c|}{\multirow{2}{*}{\textbf{Systems}}} & \multicolumn{3}{c}{\textbf{Task-wise Macro-F1 scores}}\\\cline{3-5}
& & \sent & \emot & \emotq \\ \hline
\multirow{4}{*}{\supervised} & Baseline         & 0.218          & 0.500          & 0.301          \\ \cdashline{2-5} 
& Visual BERT    & 0.320          & -          & -          \\\cdashline{2-5}
& ViLBERT    & 0.335          & -          & -          \\\cdashline{2-5}
 & Previous Best$^{\ddagger}$    & \textbf{0.355}          & \textbf{0.518}          & \textbf{0.323}          \\ \hline\hline 
\multirow{5}{*}{\selfsupervised} & SimCLR (image-only)          & 0.330          & 0.629          & 0.244          \\ \cdashline{2-5}

& VSE       &    0.248 &	0.580 &	0.292\\ \cdashline{2-5}
& VSE++$^{\dagger}$     &      0.343 &	0.675 &	0.327          \\ \cline{2-5}

 & \pienet$^{\star}$           & \textbf{0.357} & \textbf{0.755} & 0.283          \\ \cdashline{2-5} 
 & \convirt$^{\star}$          & 0.351          & 0.682          & \textbf{0.332} \\ \hline\hline
\multicolumn{2}{c|}{$\Delta_{\text{($\star$-$\dagger$)}\times 100}(\%)$} & \textcolor{blue}{$\uparrow1.37\%$} & \textcolor{blue}{$\uparrow7.93\%$}  & \textcolor{blue}{$\uparrow0.46\%$}\\ \hline
\end{tabular}
}
\caption{Comparison of \selfsupervised+\finetune\ with previous best and baseline for \memotion\ tasks. $\dagger$ represents \selfsupervised\ baseline and $\star$ is for the proposed approach and $\ddagger$ (Previous best): best scores for the corresponding tasks.}
\label{tab:memotion_base_comp}
\end{table}

\begin{figure*}[tb!]
\centering
\subfloat[{\centering \sent\ task}
\label{fig:memotion_TaskA_incremental}]{
\includegraphics[width=0.32\textwidth]{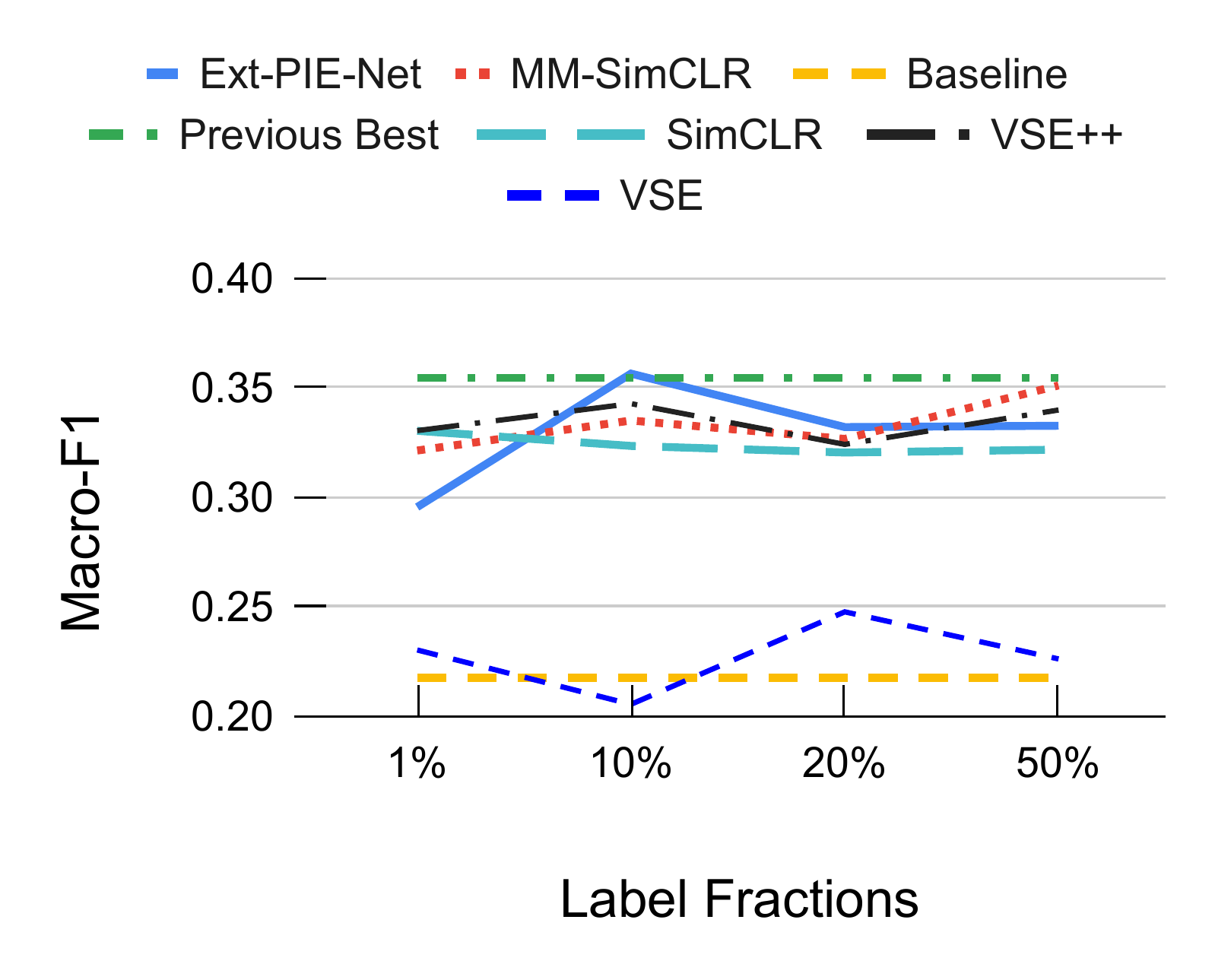}}\hspace{0.1em}
\subfloat[{\emot\ task}\label{fig:memotion_TaskB_incremental}]{
\includegraphics[width=0.32\textwidth]{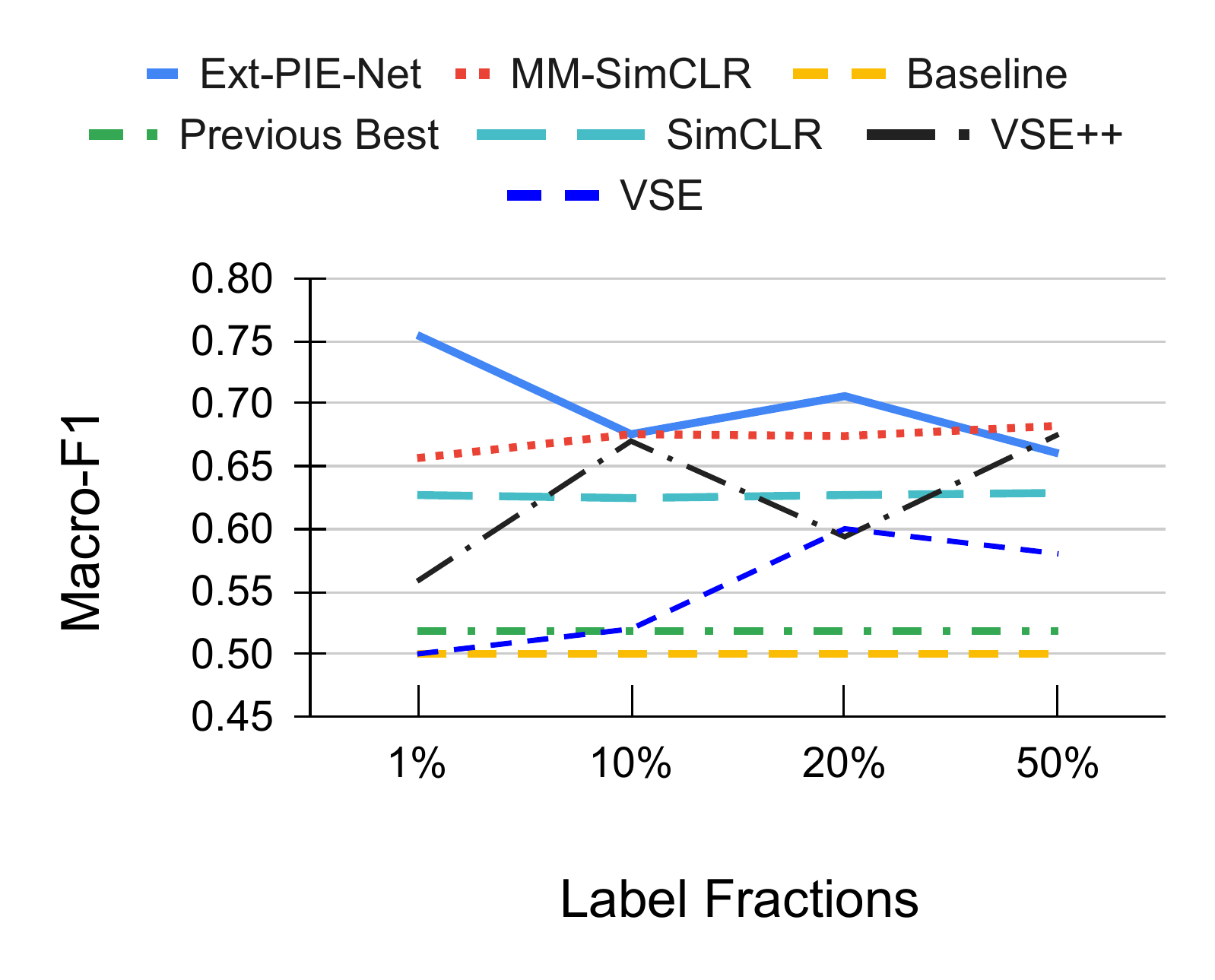}}\hspace{0.1em}
\subfloat[{\emotq\ task}\label{fig:memotion_TaskC_incremental}]{
\includegraphics[width=0.32\textwidth]{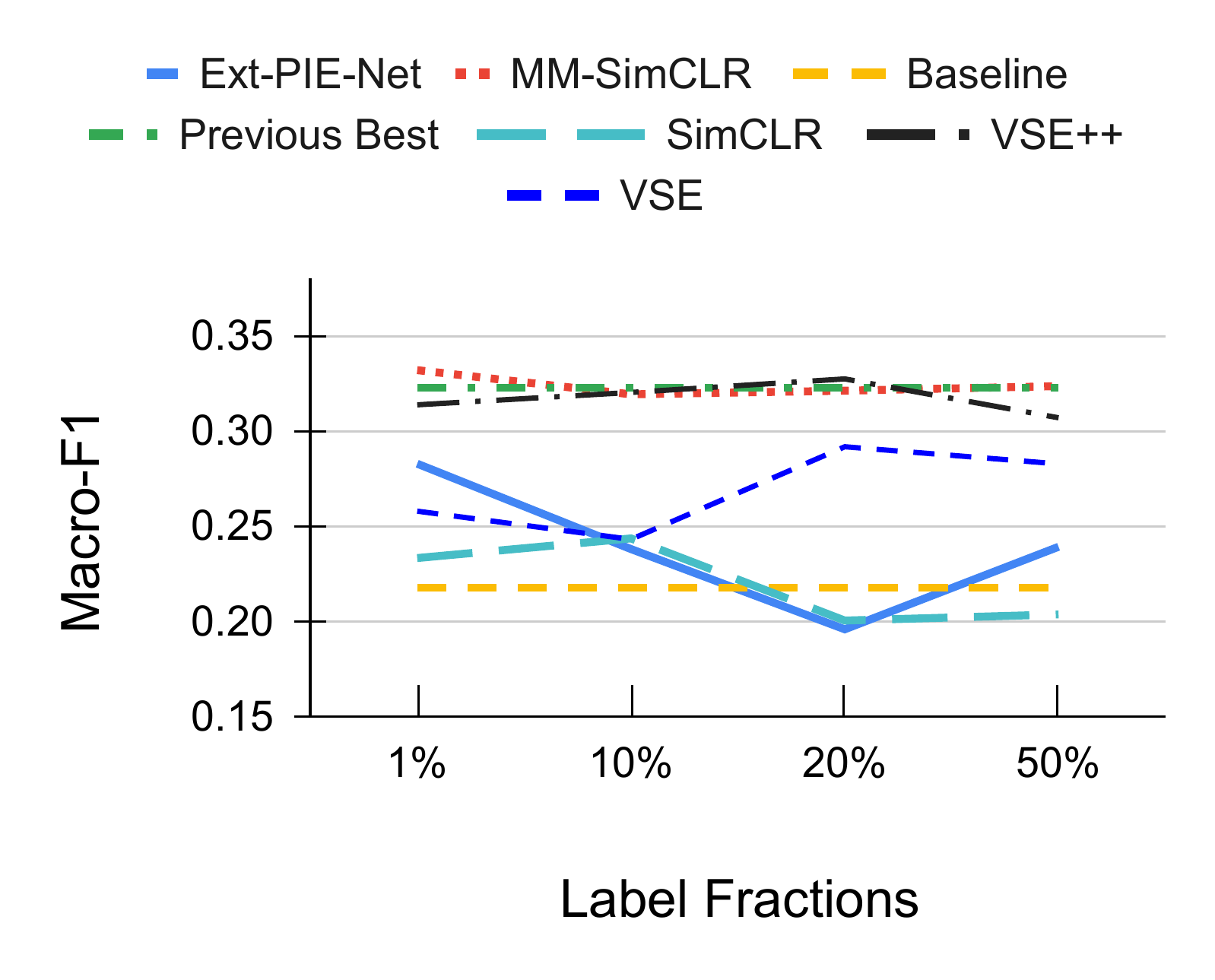}}
\caption{Comparison between the proposed method and baselines on \memotion\ tasks. X-axis signifies the incremental supervision during fine-tuning.}
\label{fig:memotion_incremental}
\end{figure*}

\subsection{Label-Efficient Training on Downstream Tasks}
We evaluate the representations learned via linear classification using a \textit{subset} of labeled samples following self-supervised pre-training to assess label efficiency during adaption. A classification head consisting of a linear layer brings the modalities into the same dimension (we use $512$). Furthermore, a shallow, fully connected network classifies the obtained multi-modal representation into target labels. We opt for the \memotion\ and \harmeme\ tasks for this paradigm. Based on the results obtained from the evaluation of self-supervision strategies, we evaluate the pre-training performance on these downstream tasks.



\textit{\underline{Results on Memotion Analysis}:} Due to the complex nature of the dataset and the tasks involved, the baselines and the leader-board for \memotion\ task \cite{sharma-etal-2020-semeval} reflect the resulting non-triviality -- with SOTA results as $0.354$, $0.518$, and $0.32$ Macro-F1 for \sent, \emot, and \emotq\ tasks,  respectively. Moreover, the complexity of the tasks can be further ascertained via the baseline's Macro-F1 scores of $0.217$, $0.500$, and $0.300$ for the three tasks -- the baseline systems are trivial early fusion (for \sent\ task), and late fusion-based (for \emot\ and \emotq\ tasks) approaches on top of CNN and RNN based image and text encoding mechanisms. The previous best systems involve a word2vec \cite{NIPS2013_9aa42b31,Mikolov2013EfficientEO} based feed-forward neural network for \sent\ \cite{keswani-etal-2020-iitk-semeval}, a multi-modal multi-tasking based setup for \emot\ \cite{vlad-etal-2020-upb}, and a feature-based ensembling approach for the \emotq\ task \cite{guo-etal-2020-guoym}. These results solicit improvement in multi-modal systems.
\begin{table}[t!]
\centering
\resizebox{0.8\columnwidth}{!}
{
\begin{tabular}{c|l|c}
\hline
\textbf{Type} & \textbf{Systems} & \textbf{Macro-F1} \\ \hline
\multirow{6}{*}{\supervised} & Late Fusion & 0.7850 \\ \cdashline{2-3} 
& Concat BERT & 0.7638 \\ \cdashline{2-3} 
& MMBT & 0.8023 \\ \cdashline{2-3} 
& ViLBERT CC                                 & 0.8603                                \\ \cdashline{2-3} 
& Visual BERT COCO                                & 0.8607                                \\ \cdashline{2-3} 
& MOMENTA & \textbf{0.8826}      \\\hline\hline 
\multirow{5}{*}{\selfsupervised} & SimCLR (image-only)                                     & 0.6328                                \\ \cdashline{2-3}

& VSE     & 0.6569                                \\\cdashline{2-3} 
& VSE++$^{\dagger}$     & 0.7912                                \\\cline{2-3} 

& \pienet     & 0.5717                                \\\cdashline{2-3} 
& \convirt$^{\star}$   & \textbf{0.8140}                                 \\ \hline\hline
\multicolumn{2}{c|}{$\Delta_{\text{($\star$-$\dagger$)}\times 100}(\%)$} & \textcolor{blue}{$\uparrow2.28\%$}\\\hline
\end{tabular}
}
\caption{Comparison of \selfsupervised+\finetune\ with previous best and baseline for \harmeme\ task.}
\label{tab:harmeme_base_comp}
\vspace{-5mm}
\end{table}

\begin{figure*}
    \centering
    \includegraphics[width=0.87\textwidth]{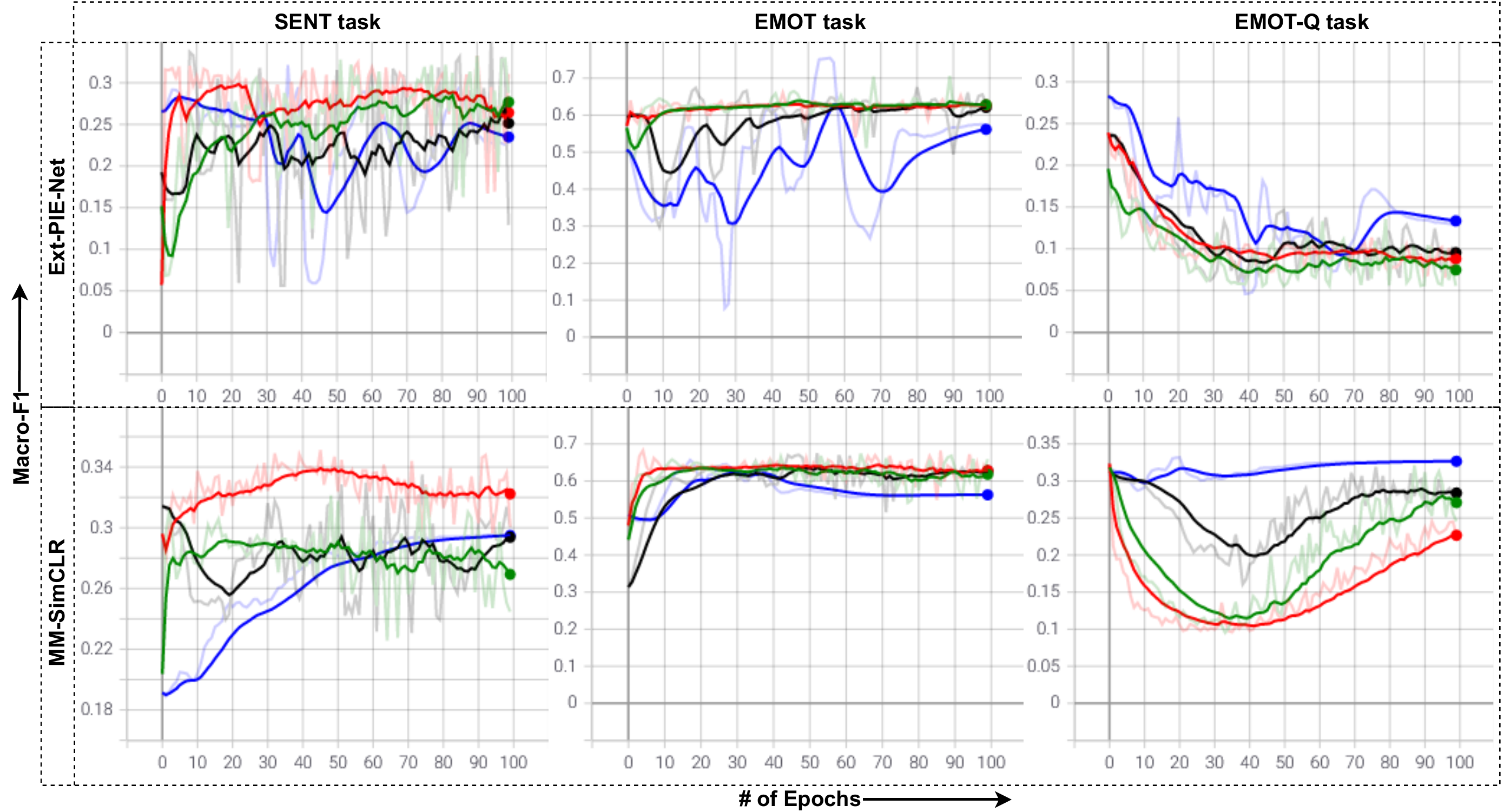}
    \caption{
    Training performance comparison for different label fractions
     \textbf{$[$\textcolor{blue}{1 \%}}
     --
     \textbf{10~\%} 
     --
     \textbf{\textcolor{darkgreen}{20~\%}}
     --
     \textbf{\textcolor{red}{50~\%}$]$} for \pienet\  (top row) and \convirt\ (bottom row) on \memotion\ tasks. Dominant curves are \textit{smoothed} depiction of the actual curves in the background.
}
\label{fig:training_curves}
\end{figure*}

We showcase the results on the same tasks by our proposed approaches in Table \ref{tab:memotion_base_comp}. \pienet\ outperforms Late-fusion baseline, Visual BERT, ViLBERT, the previous best (amongst \supervised),\ and uni-modal, multi-modal, and \convirt\ (amongst \selfsupervised) systems in the \sent\ and \emot\ tasks. It reports an improvement of $1.37\%$ in \sent\ but a significant $7.93\%$ increment over that from VSE++ (best \selfsupervised) in \emot\ at $0.3565$ and $0.7547$ Macro-F1 scores, respectively. In comparison, the performance in \emotq\ is non-convincing at $0.2827$ Macro-F1 score -- this could be due to the multi-class and multi-label nature of the task. 
Whereas, since \sent\ and \emot\ tasks are formulated by aggregating data samples for the higher level of categorical consideration, they are relatively complex due to the resulting data imbalance. Although \convirt\ performs better on \emotq\ task and overall, at-par or better than the baseline, it still lags by a small margin for \sent\ task and significantly for Task B compared to \pienet. Also, \pienet\ setup has a relatively more significant number of trainable parameters than \convirt, facilitating better modeling capacity for \sent\ and \emot\ tasks. Conversely, \convirt\ performs better on \emotq\ task due to better compatibility of the modeling capacity and task. The overall results signify the efficacy of proposed \selfsupervised\ strategies on complex downstream multi-modal tasks. These results highlight the task-specific peculiarities that modeling needs to factor in for optimal performance.


\textit{\underline{Results on Harmful Memes}:} The transferability of the representations learned through pre-training is examined by fine-tuning on another meme dataset, i.e., \harmp. 
We report the results in Table \ref{tab:harmeme_base_comp}. The fully supervised models, such as VilBERT CC \cite{pramanick-etal-2021-momenta-multimodal}, Visual BERT COCO \cite{pramanick-etal-2021-momenta-multimodal}, and MOMENTA \cite{pramanick-etal-2021-momenta-multimodal}, obtain Macro-F1 scores of $0.8603$, $0.8607$, and $0.8826$, respectively.
In comparison, \convirt\ in a label-efficient setup records a convincing performance of $0.8140$ Macro-F1.
One of our proposed approaches \pienet\ performs poorly with $0.5717$ F1 against an impressive F1 score of $0.8140$ by \convirt.\ Like its performance on \memotion\ task, \convirt\ is observed to perform better on a relatively more straightforward \harmeme\ task. Even though \convirt\ lags behind by $4.6\%$ from strong \supervised\ baselines ViLBERT CC and Visual BERT COCO, and MOMENTA by $7.02\%$, it distinctly outperforms other competitive multi-modal baselines (supervised) like Late Fusion, Concat BERT and MMBT by $2.9\%$, $5.02\%$ and $1.87\%$, respectively. \convirt\ also leads SimCLR ($0.6328$) by $18.12\%$, and \selfsupervised\ multi-modal baselines VSE ($0.6569$), VSE++ ($0.7912$) and \pienet\ ($0.5717$) by $15.71\%$, $2.28\%$ and $24.2\%$, respectively on the \harmeme\ task.

It is also worth highlighting that the performances of strong multi-modal models like Visual BERT and ViLBERT can be inconsistent, depending upon the task being addressed. This is primarily due to the fact that the corresponding pre-training involved leverages strong visual-linguistic grounding, which based on downstream task complexity, can give varying results as observed for \memotion\ (c.f. Table. \ref{tab:memotion_base_comp}) and \harmeme\ (c.f. Table \ref{tab:harmeme_base_comp}). This suggests the scope of enhancement towards the pre-training objectives and frameworks within the existing multi-modal systems.

\section{Impact of Label-Efficient Supervision During Fine-tuning}
Towards assessing the label-efficient setup, we compare the performances over incremental supervision. We also analyze their temporal training behavior.

As can be observed from Fig. \ref{fig:memotion_TaskA_incremental}, \pienet\ converges efficiently to $0.3565$ F1 score with just 10$\%$ ($600$) training samples, as compared to \convirt\ which converges to $0.3511$ F1 score after learning from 50$\%$ (3000) of the labeled samples. This highlights the capacity of a sophisticated \selfsupervised\ regime to learn better representations for a complex setup for the \sent\ task compared to a slightly simpler model \convirt. 
A similar pattern can be observed for \emot\ task in Fig. \ref{fig:memotion_TaskB_incremental}. \pienet\ is observed to achieve an overall better F1 score of $0.7547$, which is better than \convirt\ and outperforms all other results.

Although the optimal performance of SimCLR is reasonably at-par or even better for \sent\ and \emot\ tasks compared to the baseline and the previous best results, there is barely any active convergence visible within the plots depicted in Fig. \ref{fig:memotion_incremental} for it. This is obvious considering the incomplete information that an image-only based uni-modal system would learn for the downstream task. VSE is observed to yield $3.02\%$ and $7.98\%$ improvement over the \supervised\ baseline. Still, it fails to register an impressive performance compared to the increment of $12.52\%$ and $17.52\%$ for the two tasks, respectively, by VSE++.

\begin{figure}
    \centering
    \includegraphics[width=\columnwidth]{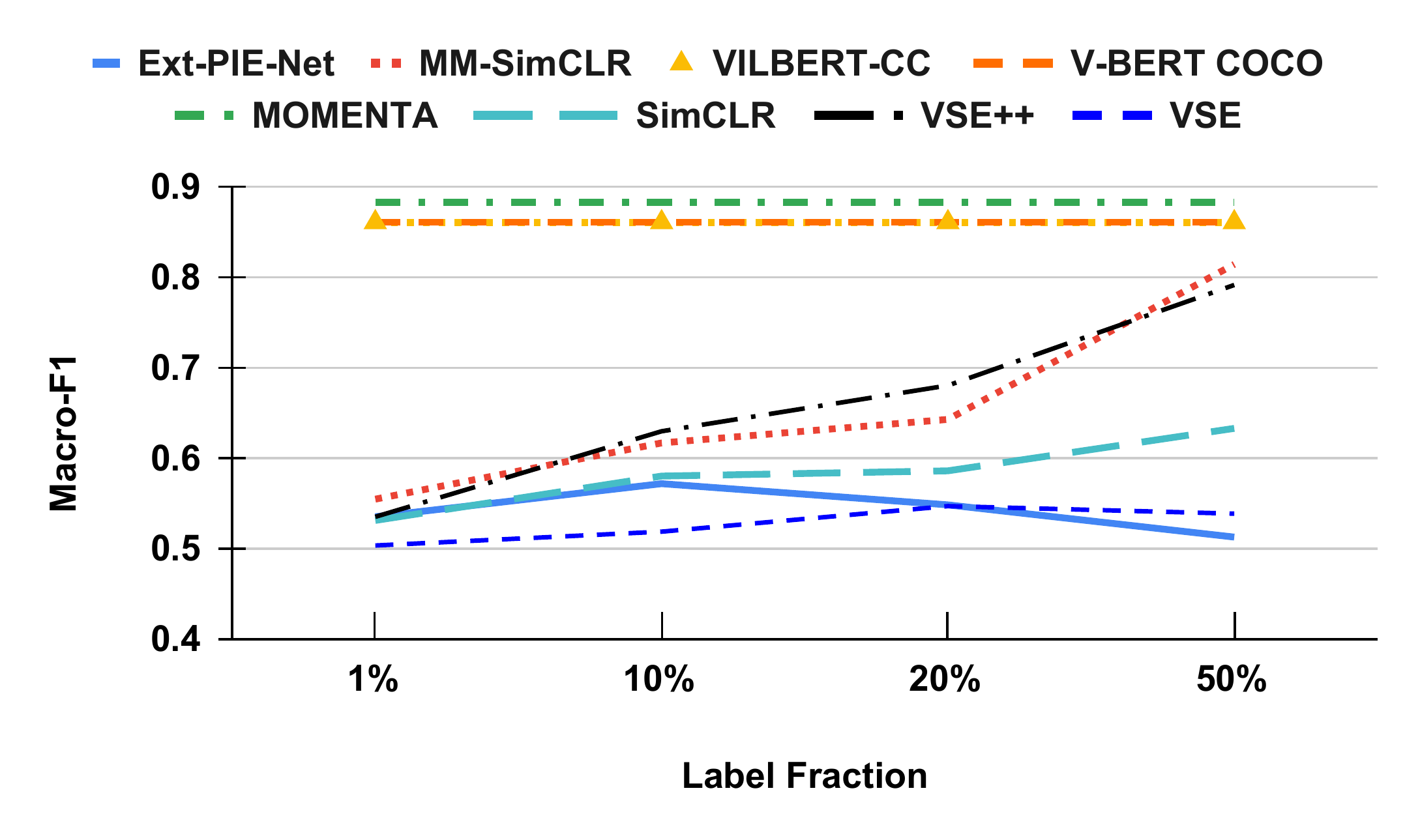}
    \caption{Comparison b/w the proposed method and baselines on \harmeme\ tasks. X-axis signifies the incremental supervision during fine-tuning.}
    \label{fig:harmeme_incremental_comp}
\end{figure}

These observations can also be correlated with the training performance (c.f. Fig. \ref{fig:training_curves}), wherein the performance curves are depicted for a total of 100 epochs across four different label-efficiency scenarios. For primary assessment, we showcase \textit{smoothed} curves overlaid on \textit{unsmoothed} ones towards observing global and local trends. \footnote{For further reference, \textit{unsmoothed} training curves are also included and discussed separately in App. \ref{app:curves}.} 

Fig. \ref{fig:training_curves} presents a clear depiction of progressive learning for all the supervision configurations evaluated in case of \pienet\ for the \sent\ and \emot\ tasks (c.f. Fig. \ref{fig:training_curves}) is given. On the other hand, the training curves for \convirt\ show saturated learning for tasks \sent\ and \emot\ respectively (c.f. Fig. \ref{fig:training_curves}). 


Delineating on the performance trend observed in the \emotq\ task earlier, neither \pienet\ nor SimCLR shows definite convergence, as we consider the incremental supervision depicted in Fig. \ref{fig:memotion_TaskC_incremental}. Whereas, \convirt\ is observed to show stable, yet non-incremental growth in performance reporting the best overall F1 score of $0.3318$ (c.f. Table \ref{tab:memotion_base_comp}). This task entails a relatively balanced training set \cite{sharma-etal-2020-semeval}, and \convirt\ is observed to offer just the required simplicity for solving such a task. The training characteristics observed for this task, are found to be contrasting for \pienet\ and {\convirt} (c.f. Fig. \ref{fig:training_curves}, last figures from \textit{first and second rows, respectively}). 
\convirt\ indicates overall progressive learning. On the other hand, \pienet\ depicts a consistently regressive trend. This corroborates the optimal convergence demonstrated by a simple multi-modal contrastive loss-based self-supervision for a more straightforward task formulation.

For \harmeme\ task, the incremental supervision (c.f. Fig. \ref{fig:harmeme_incremental_comp}) exhibits incremental performance with the increase in the amount of supervision during fine-tuning. Notably, the final F1 score of $0.814$ obtained by the \convirt\ model is on just 50 $\%$ (1510) of the actual training set. This demonstrates the efficacy and generalizability of the pre-training via strategies adopted in this work.
Also, the progressive convergence observed at $50\%$ supervision, as shown in Fig. \ref{fig:harmeme_incremental_comp} for \convirt,\ demonstrates the generalizability of the proposed approach. This also suggests the importance of having smaller architectures with sophisticated fusion strategies to solve the task at hand effectively.

\section{Discussion}
The observations made from the results obtained for the downstream evaluation suggest interesting trends. Since \memotion\ dataset involves multi-class, multi-label and multi-level hierarchical granularity due to the natural distribution of such realistic dataset, either ensembling-based approaches are observed to yield better results or, there are strong variations observed in the performance trends across the three \memotion\ tasks \cite{sharma-etal-2020-semeval}. The results reported as part of Table \ref{tab:res_ssl_base}, \ref{tab:memotion_base_comp} and \ref{tab:harmeme_base_comp} exhibit insights correlating the task complexity with that of the modelling solutions required. This is further corroborated by the results on \harmeme\ task. To this end, we have highlighted the performances and drawn comparisons for two models that we empirically examined as part of this investigation.

\section{Conclusion}

This work empirically examined various self-supervision strategies to learn effective representations that help solve multiple multi-modal downstream tasks in a label-efficient setting. We propose two strategies for this -- (i) \convirt:\ a multi-modal contrastive loss formulation that factors in the loss terms for image modality and the multi-modality in a joint manner, and (ii) \pienet:\ a joint formulation of weighted modality-specific hinge loss terms, combined with the contrastive loss that is computed between a pair of representations, obtained using symmetric multi-modal fusion. Extensive analysis over 2 datasets and 5 tasks demonstrate how domain-aware self-supervised pre-training, using a multi-modal dataset, that can be directly obtained from the wild (WWW) in raw form, can be leveraged to perform label-efficient multi-modal adaptation, leading to competitive, even superior performance gains for some scenarios.

The performances observed for the proposed methods indicate \textit{task-dependent} efficacies. \convirt\ being a lighter model is observed to perform better on \emotq\ and \harmeme\ tasks, having a lower level of granularity to be modeled. Whereas \pienet\ performs better on \sent\ and \emot\ tasks, which require modeling a higher abstraction level for the target categories. Despite exhibiting interesting performance within label-efficient evaluation settings, the objectives addressed in this work can further benefit from extensive analysis and evaluation towards obtaining a broader understanding of the generalizability of the proposed methodology. 

\section*{Acknowledgments}
The work was supported by Wipro research grant.

\bibliography{anthology,custom}

\appendix
\clearpage
\begin{table}[h!]
\centering
\resizebox{\columnwidth}{!}{%
\begin{tabular}{c|c|c|c|c|c|c} 
\hline
\textbf{Type}                    & \textbf{Name} & \textbf{BS}  & \textbf{Epochs}      & \textbf{LR}  & \textbf{Image Encoder}     & \textbf{Text Encoder}                     \\ 
\hline
\multirow{5}{*}{SSL} & SimCLR        & \multirow{5}{*}{32}  & 150                  & 0.1                     & ResNet-50                  & -                                         \\ 
\cdashline{2-2}\cdashline{4-7}
                                 & VSE++         &                      & \multirow{4}{*}{100} & \multirow{4}{*}{0.0001} & \multirow{4}{*}{ResNet-18} & \multirow{4}{*}{distilbert-base-uncased}  \\ 
\cdashline{2-2}
                                 & Mod. SimCLR     &                      &                      &                         &                            &                                           \\ 
\cdashline{2-2}
                                 & MM-SimCLR     &                      &                      &                         &                            &                                           \\ 
\cdashline{2-2}
                                 & Ext-PIE-Net   &                      &                      &                         &                            &                                           \\ 
\hline\hline
\multirow{3}{*}{SL}      & SimCLR        & 512                  & \multirow{3}{*}{100} & 0.0001                  & ResNet-50                  & -                                         \\ 
\cdashline{2-3}\cdashline{5-7}
                                 & MM-SimCLR     & \multirow{2}{*}{256} &                      & \multirow{2}{*}{0.0005} & \multirow{2}{*}{ResNet-18} & \multirow{2}{*}{distilbert-base-uncased}  \\ 
\cdashline{2-2}
                                 & Ext-PIE-Net   &                      &                      &                         &                            &                                           \\
\hline
\end{tabular}
}
\caption{Hyperparameter values for the experiments.}
\label{tab:hyperparams}
\end{table}
\section{Experimental setup and Hyperparameters:} 
\label{sec:hyper}
We train all our experiments using Pytorch on an NVIDIA Tesla P$4$ with $8$ GB dedicated memory. We use VISSL, an open-source library \cite{goyal2021vissl} to evaluate SimCLR, a uni-modal image-only setup for memes. For the multi-modal setups, we initialize the networks with weights of pre-trained models available for image encoders with PyTorch library and the text models with weights available from $\texttt{transformers}$ package from hugging face library\footnote{https://huggingface.co}. 

\begin{table*}[t!]
\centering
\resizebox{\textwidth}{!}{
\begin{tabular}{c|c|c|c|c|c|c|c|c|c|c|c} 
\hline
\multicolumn{4}{c|}{\textbf{\mmhs}}                                            & \multicolumn{4}{c|}{\textbf{\facebook}}                                                & \multicolumn{4}{c}{\textbf{\harmp}}                                              \\ 
\hline
\multicolumn{2}{c|}{\textbf{Hateful}} & \multicolumn{2}{c|}{\textbf{Not-hateful}} & \multicolumn{2}{c|}{\textbf{Hateful}} & \multicolumn{2}{c|}{\textbf{Not-hateful}} & \multicolumn{2}{c|}{\textbf{Harmful}} & \multicolumn{2}{c}{\textbf{Not-harmful}}  \\ 
\hline
\textit{Word} & \textit{tf-idf score}  & \textit{Word} & \textit{tf-idf score}     & \textit{Word} & \textit{tf-idf score} & \textit{Word} & \textit{tf-idf score}     & \textit{Word} & \textit{tf-idf score} & \textit{Word} & \textit{tf-idf score} \\ 
\hline
faggot & 0.0441 & redneck & 0.0099 & black & 0.0433 & like & 0.0337 & photoshopped  & 0.0589 & party & 0.02514 \\ 
\hline
cunt & 0.0364 & love & 0.0098 & white & 0.0378 & day & 0.018 & married & 0.0343 & debate & 0.0151 \\ 
\hline
nigger & 0.0346 & happy & 0.0081 & muslim & 0.0321 & got & 0.0174 & joe & 0.0309 & president & 0.0139 \\ 
\hline
retarded & 0.0306 & good & 0.0074 & jews & 0.0239 & time & 0.0172 & trump & 0.0249 & democratic & 0.0111 \\ 
\hline
trash & 0.0214 & hillbilly & 0.0071 & kill & 0.0223 & love & 0.0138 & nazis & 0.0241 & green & 0.0086 \\
\hline
\end{tabular}
}
\vspace{-3mm}
\caption{The top-5 most frequent words and their tf-idf scores in each class.}
\label{tab:tfidf_mmhsfbhm}
\end{table*} 

The image encoder is a ResNet-18 \cite{He2016DeepRL} architecture and the text encoder is a \texttt{distilbert-base-uncased} in all our multi-modal experiments. After self-supervised pre-training, we freeze the text and image encoder weights and discard the projection heads attached. As part of the classification head, a new set of layers are added to perform supervised learning using fewer labeled samples. We initialize the layers using Xavier initialization \cite{pmlr-v9-glorot10a} and set the bias to zero. We train all the models using the Adam optimizer \cite{KingmaB14} and a cross-entropy loss as the objective function for supervision for all the tasks evaluated in this work. We perform multi-modal self-supervision experiments keeping a batch size of $32$ for $100$ epochs at a learning rate of $0.0001$. The SimCLR experiment in self-supervision is carried out for $150$ epochs with a batch size of $32$ and a learning rate of $0.1$ using a ResNet-$50$ backbone. The encoder weights are frozen during the label-efficient training, and the classification heads are used, allowing 256 batch-size in multi-modal experiments and $512$ for uni-modal SimCLR experiment. The SimCLR-based label-efficient setup is trained with $0.0001$ learning rate, while the other multi-modal experiments are trained with $0.0005$ learning rate. We also present these details in Table \ref{tab:hyperparams}.

\section{Statistical Analysis of Datasets}
\label{app:datanalysis}
The datasets used in this work have been either created synthetically using specific hate topics or downloaded from social media platforms using generic and domain-specific hate keywords \cite{kiela2020hateful,Gomez2020ExploringHS,pramanick-etal-2021-momenta-multimodal}. The top-5 hate and non-hate keywords ranked as per the tf-idf scores of their occurrences within the accompanying texts are shown in Table \ref{tab:tfidf_mmhsfbhm}. This table shows that the hateful lexicon for \mmhs\ represents extreme urban parlance, depicting realistic social media communication, whereas in the \facebook\ dataset, hate keywords are canonical and topic-oriented. To counter the potential keyword bias within the datasets, the categorical representation of these keywords was explicitly balanced by introducing confounders or considering contrastive examples for the exact hate keywords.

\begin{figure*}[t!]
\centering
\subfloat[{\centering \mmhs}\label{fig:hist_mmhs}]{
\includegraphics[width=0.31\textwidth]{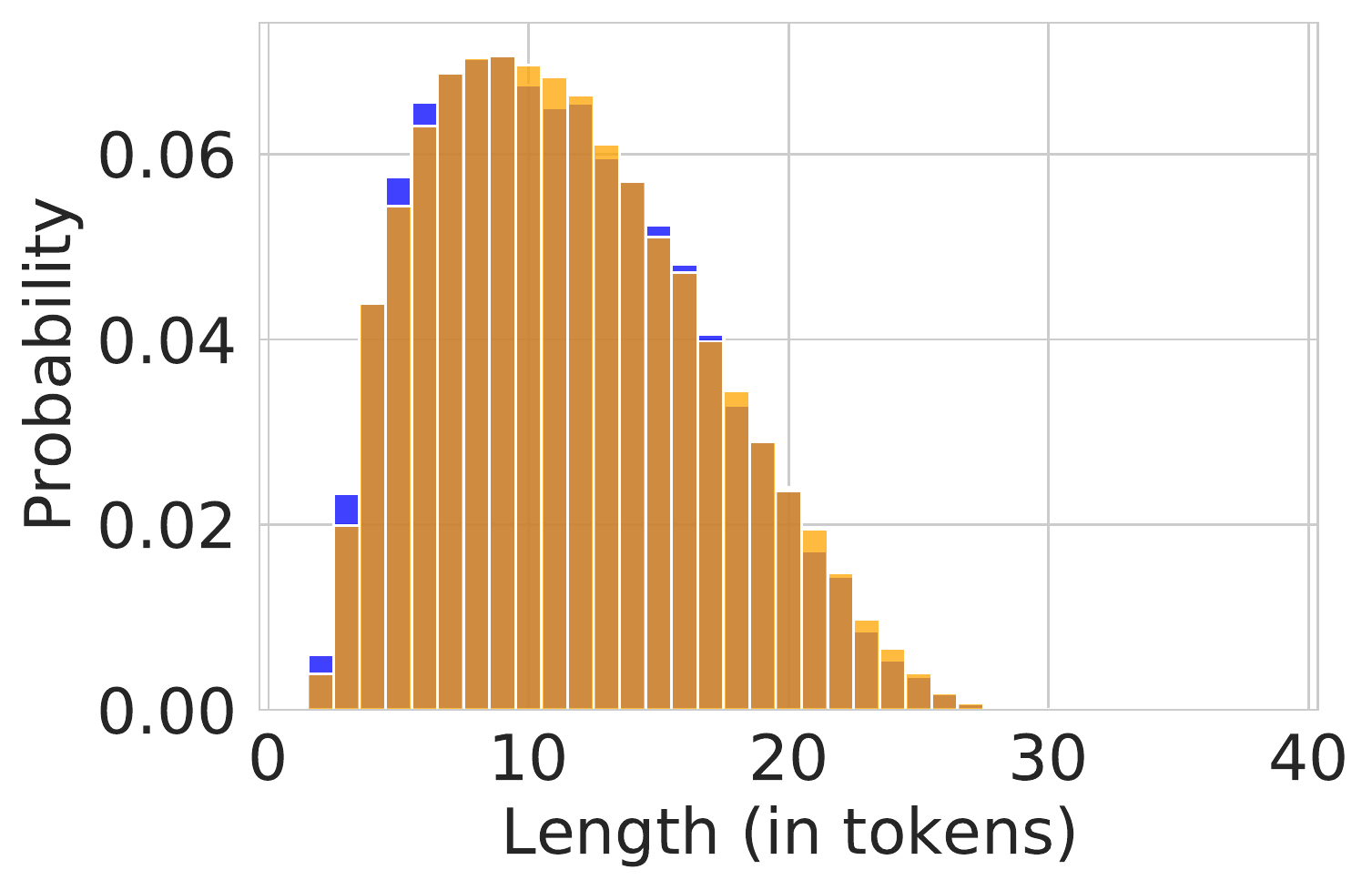}}\hspace{0.1em}
\subfloat[{\facebook}\label{fig:hist_fbhm}]{
\includegraphics[width=0.31\textwidth]{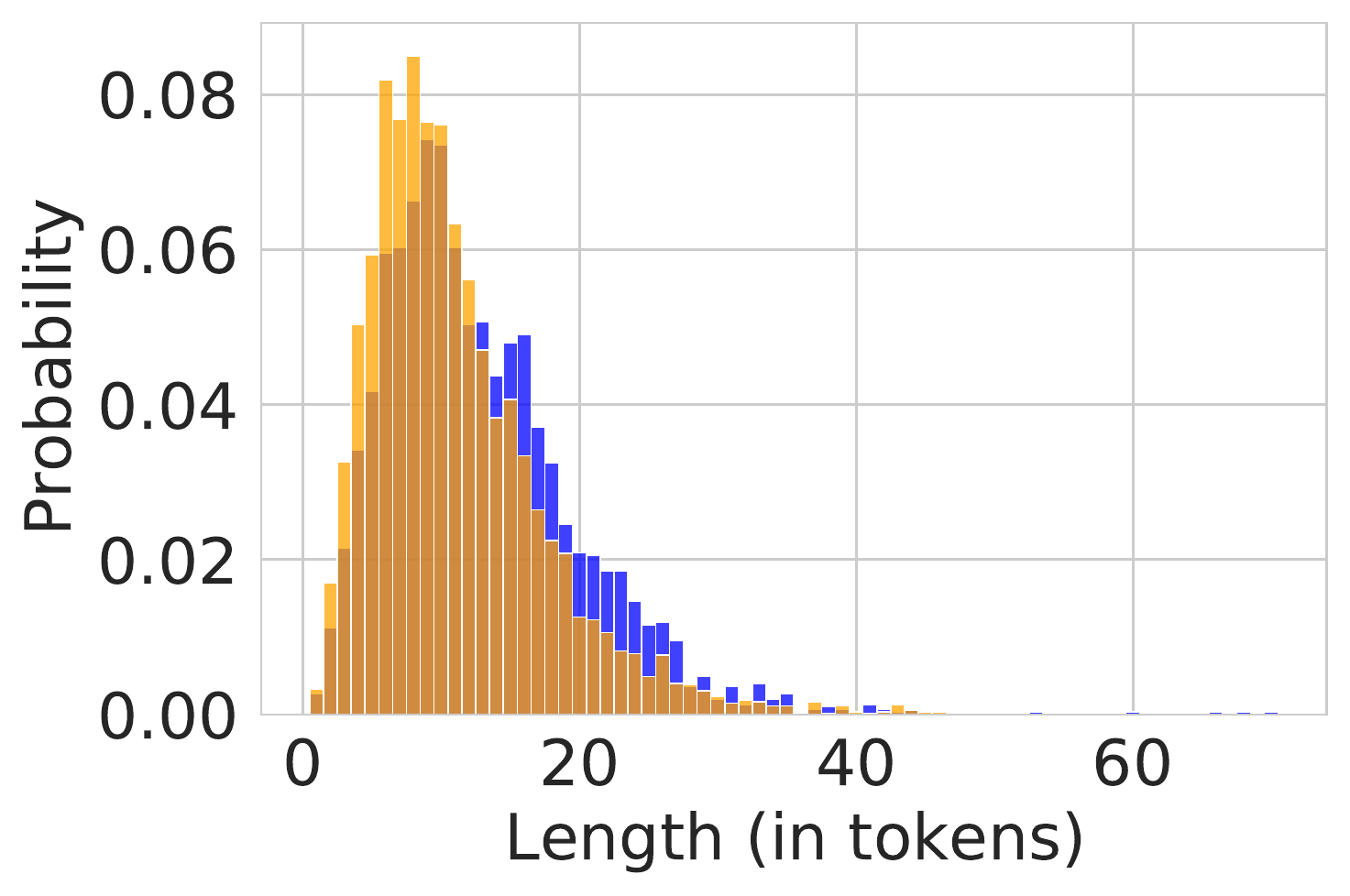}}\hspace{0.1em}
\subfloat[{\centering \harmp}\label{fig:hist_harmeme}]{
\includegraphics[width=0.31\textwidth]{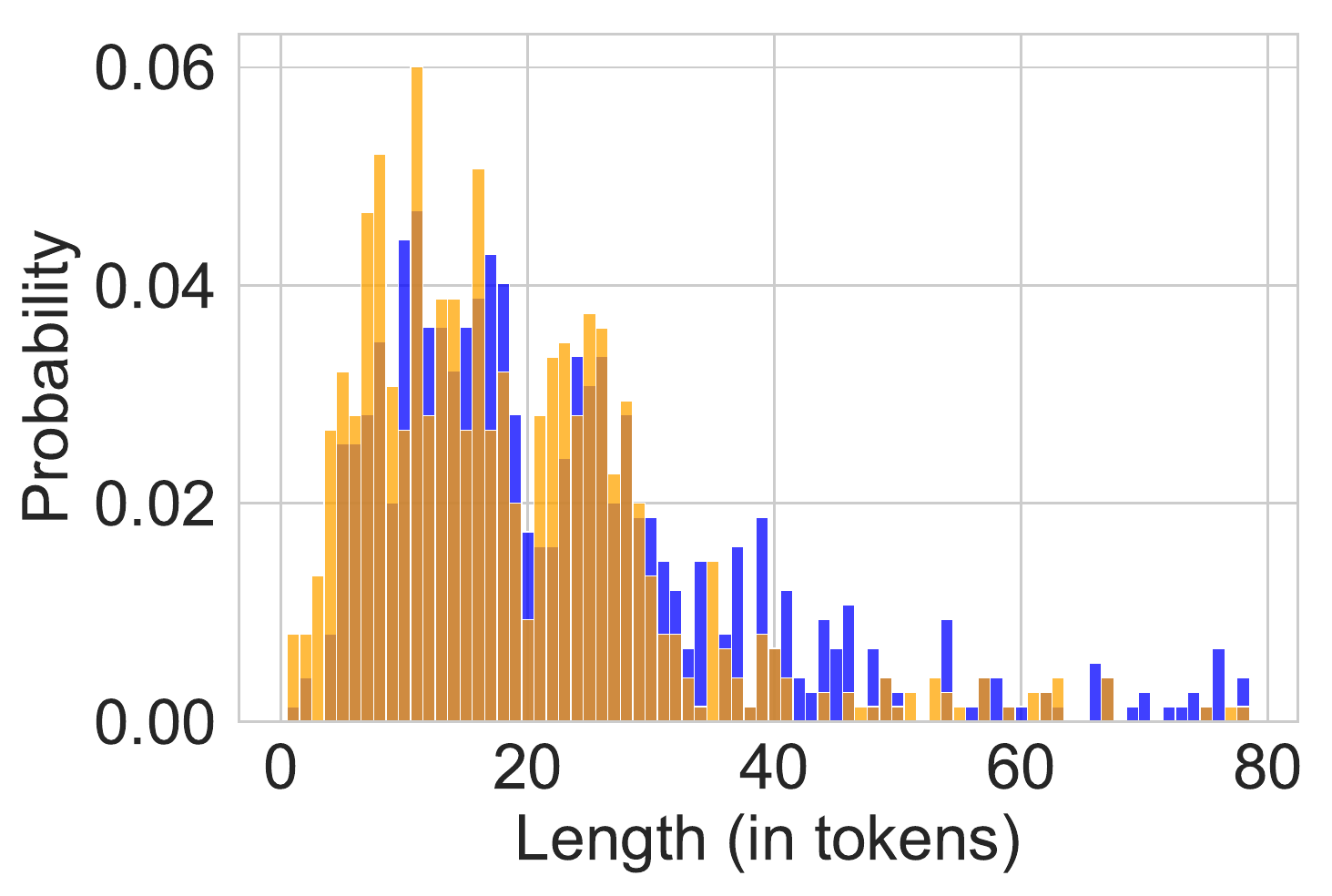}}
\vspace{-3mm}
\caption{Distributions of the text's length. Blue: Hateful/Harmful; Orange: Not-hateful/harmful.}
\label{fig:hist_all}
\end{figure*}

The accompanying texts from all datasets used have a mean length of $8$ (c.f. Fig. \ref{fig:hist_all}). The distribution observed for \mmhs\ in Fig. \ref{fig:hist_mmhs} is almost uniform, with most of the posts having lengths of less than 30 words, primarily due to the 280-character limit on tweets. \facebook,\ on the other hand, is created with reasonable variation, having examples with lengths greater than 30 as well. Their confounding effect is also clearly visible within these histogram plots, where hateful content with larger corresponding text could also be present in some samples (Fig. \ref{fig:hist_fbhm}), as against the general trend where the variation in the length is confined. Finally, \harmp\ reflects the distribution of the accompanying textual contents over social media. Hence the variation depicted in Fig. \ref{fig:hist_harmeme}.

\begin{figure*}[t!]
    \centering
    \includegraphics[width=0.9\textwidth]{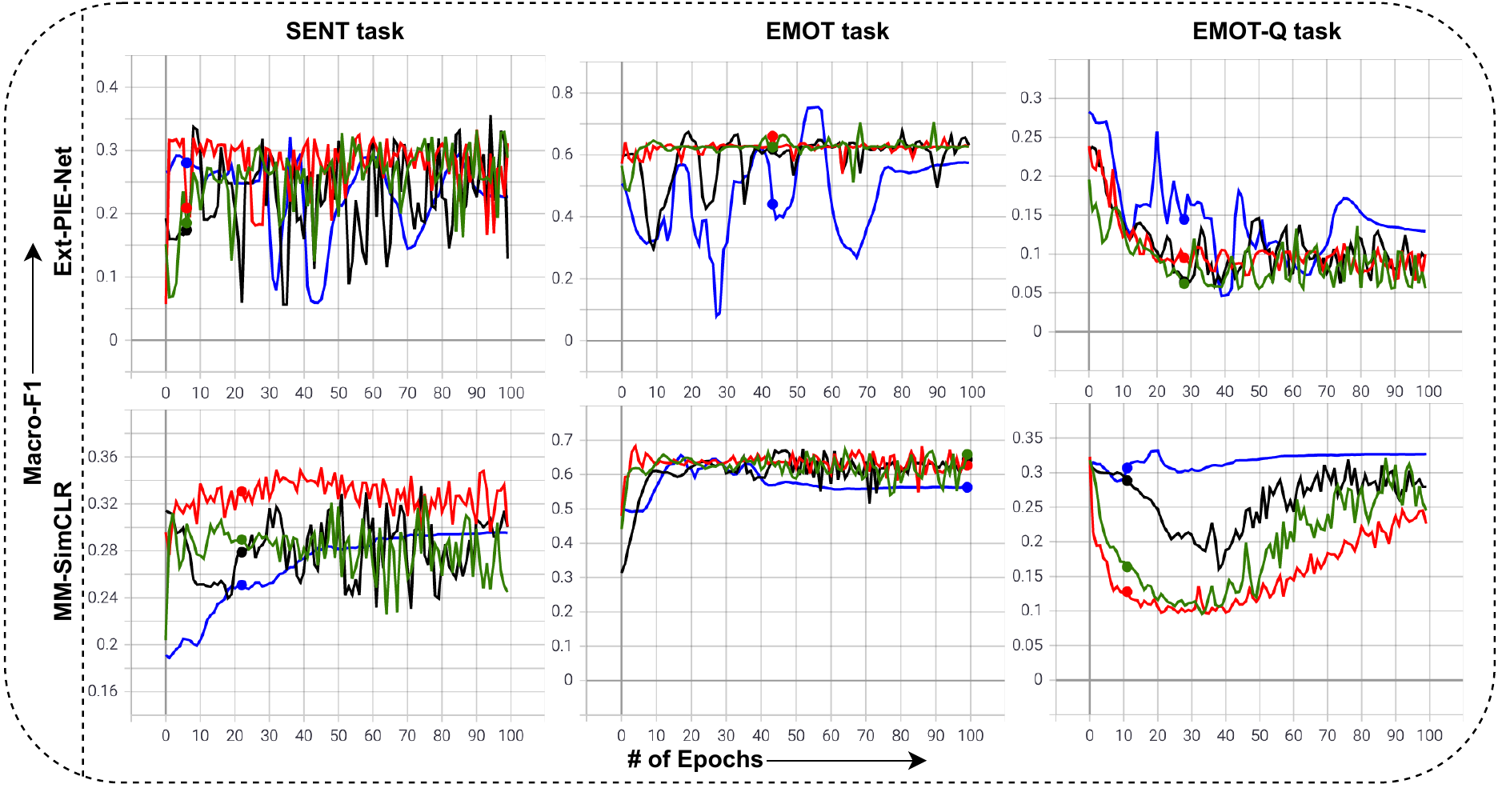}
    \caption{
    Training performance comparison (\textit{unsmoothed}) for different label fractions
     \textbf{$[$\textcolor{blue}{1 \%}}
     --
     \textbf{10~\%} 
     --
     \textbf{\textcolor{darkgreen}{20~\%}}
     --
     \textbf{\textcolor{red}{50~\%}$]$} for \pienet\  (top row) and \convirt\ (bottom row) on \memotion\ tasks.
}
\label{fig:unsmoothed_curves}
\end{figure*}

\section{Training Characteristics}
\label{app:curves}
The \textit{unsmoothed} training curves, depicted in Fig. \ref{fig:unsmoothed_curves} reflects the trends observed with the \textit{smoothed} depiction in Fig. \ref{fig:training_curves}. Besides significant fluctuations within the training curves across tasks, especially for \sent\ and \emotq\ tasks, subtle temporal trends can be inferred. There is a gradual enhancement in the performances observed within early epochs ($<$60) for both \sent\ and \emot\ tasks, for both \pienet\ and \convirt, with \pienet\ registering the best macro-f1, along with significant variation. But overall, the performances are reasonably similar. For \sent\ task, \pienet\ showcases consistent growth in the macro-f1 score for all the label-configuration scenarios. In contrast, \convirt\ showcases progress for scenarios involving $1\%$ and $50\%$ labeled samples only. On the other hand, for \emotq\ task, \convirt\ is observed to exhibit better convergence after $30^{th}$ epoch, as against that by \pienet, across label-configurations, suggesting better training behavior (c.f. Fig. \ref{fig:unsmoothed_curves}).

\section{Ethics and Broader Impact}
\label{app:ethics}
 
\paragraph{User Privacy.} 
The meme content and the associated information do not include any personal information. Issues related to copyright are addressed as part of the dataset source.  


\paragraph{Biases.}

Any biases found in the datasets \cite{Gomez2020ExploringHS,kiela2020hateful,pramanick-etal-2021-momenta-multimodal} leveraged in this work are presumed to be unintentional, as per the attributions made in the respective sources, and we do not intend to cause harm to any group or individual. We acknowledge that detecting emotions and harmfulness can be subjective, and thus it is inevitable that there would be biases in gold-labeled data or the label distribution. The primary aim of this work is to contribute with a novel multi-modal framework that helps perform downstream-related tasks, utilizing the representations learned via self-supervised learning.

\paragraph{Misuse Potential.}
We find that the datasets used in this work can be potentially used for ill-intended purposes, like biased targeting of individuals/communities/organizations, etc., that may or may not be related to demographics and other information within the text. Any research activity would require intervention with human moderation to ensure this does not occur.

\paragraph{Intended Use.}

We use the existing dataset in our work in line with the intended usage prescribed by its creators and solely for research purposes. This applies in its entirety to its further use as well. We commit to releasing our dataset, aiming to encourage research in studying harmful targeting in memes on the web. We distribute the dataset for research purposes only, without a license for commercial use. We believe that it represents a valuable resource when used appropriately.

\paragraph{Environmental Impact.}

Finally, due to the requirement of GPUs/TPUs, large-scale Transformers require many computations, contributing to global warming \cite{strubell2019energy}. However, in our case, we do not train such models from scratch; instead, we fine-tune them on relatively small datasets. Moreover, running on a CPU for inference, once the model has been fine-tuned, is perfectly feasible, and CPUs contribute much less to global warming.

\end{document}